\documentclass[runningheads,a4paper]{llncs}

\usepackage{amssymb}
\usepackage{amsmath}
\usepackage{comment}
\usepackage{graphicx}
\usepackage{tikz}
\usepackage{syntax}
\usepackage{import}
\usetikzlibrary{arrows,shapes,snakes,automata,backgrounds,petri,shadows,positioning}
\usepackage{float}
\usepackage{booktabs} 
\usepackage{subcaption}
\usepackage{marvosym} 

\usepackage[formats]{listings}
\lstdefineformat{mine}{~=\( \sim \)}
\usepackage[obeyspaces]{url}
\urldef{\mailsa}\path|g.sileno@uva.nl|     
\newcommand{\keywords}[1]{\par\addvspace\baselineskip 
	\noindent\keywordname\enspace\ignorespaces#1}

\newcommand{\tuple}[1]{\ensuremath{\left \langle #1 \right \rangle }} 
\DeclareMathOperator*{\mynot}{not}

\makeatletter 
\renewcommand*{\p@section}{\S\,} 
\renewcommand*{\p@subsection}{\S\,} 
\renewcommand*{\p@subsubsection}{\S\,} 
\makeatother{} 

\begin{document} 
	
	\mainmatter  
	
	\title{Operationalizing Declarative and Procedural Knowledge: a Benchmark on Logic Programming Petri Nets (LPPNs)} 
	\titlerunning{Logic Programming Petri Nets: a Benchmark} 
	
	%
	%
	\author{Giovanni Sileno$^{1}$}%
	\authorrunning{Logic Programming Petri Nets: a Benchmark}
	
	\institute{Informatics Institute, University of Amsterdam, the Netherlands\\
		\mailsa} 
	
	%
	%
	
	\toctitle{Logic Programming Petri Nets: a benchmark} 
	\tocauthor{Giovanni Sileno} 
	\maketitle 

\begin{abstract}
Modelling, specifying and reasoning about complex systems requires to process in an integrated fashion declarative and procedural aspects of the target domain. The paper reports on an experiment conducted with a propositional version of Logic Programming Petri Nets (LPPNs), a notation extending Petri Nets with logic programming constructs. Two semantics are presented: a denotational semantics that fully maps the notation to ASP via Event Calculus; and a hybrid operational semantics that process separately the causal mechanisms via Petri nets, and the constraints associated to objects and to events via Answer Set Programming (ASP). These two alternative specifications enable an empirical evaluation in terms of computational efficiency. Experimental results show that the hybrid semantics is more efficient w.r.t. sequences, whereas the two semantics follows the same behaviour w.r.t. branchings (although the denotational one performs better in absolute terms).
\end{abstract}


\section{Introduction}
A proper treatment of cases or scenarios is based on two requirements: on the one hand, to capture and adequately process the symbolic entities used to \emph{represent} the target system: instances, classes, interrelationships forming a local ontology relevant to the domain in focus; on the other hand, to \emph{reproduce}---by means of elements modelling causal mechanisms, processes, courses of actions, etc.---the same dynamics exhibited by the target system. 

Consider for example this case: ``\textit{While John was walking his dog, the dog ate Paul's flowers}.'' This event description is not sufficient for entailing that John is responsible to pay Paul for what happened, as typically this is entailed on the basis of norms as ``\textit{The owner of an animal has to pay for the damages it produces.'}'. 
However, even this addition lacks crucial connections between the conceptual domains of the case and the one of the norm, like ``\textit{dogs are animals}'', ``\textit{eating an object destroys the object}'', ``\textit{destruction is damage}'', etc. 

These various elements exhibit two perspectives on knowledge: a \textit{declarative} perceptive, concerning \emph{objects} (physical, mental, institutional) and their logical relationships---both reified as symbols---; and a \textit{procedural} perceptive, concerning patterns of events/actions, mechanisms, or \emph{processes} (involving objects). Formal logic is the prototypical domain concerned with the first perspective, just as process modeling focuses on the second. Unfortunately, methodologies associated with one of the two aspects generally have a limited treatment of the other component, and they require specific mediating machinery to deal with. For instance, if you want to make a certain outcome impossible in a procedural model you need to add conditions that disable all transitions that might produce that outcome. If you want to represent a transition in a declarative way, a typical approach is to consider \emph{snapshots} of the arrangements holding before and after the transition---possibly labeled with a sort of timestamp. This is essentially the principle behind \emph{situation calculus} \cite{McCarthy1969,Reiter2001}, \emph{event calculus} \cite{Kowalski1986,Shanahan1999}, and \emph{fluent calculus} \cite{Thielscher1999}: using appropriate axioms, you can create and reason about the relations between these snapshots in a way such that they are compatible to the causal relationships between the moments they refer to.  

Rather than trying to project one dimension on the other, an alternative tradition in AI and logic proposes to consider \textit{causality} as a primitive notion. This approach is for instance behind the idea of all \emph{Action languages} \cite{Gelfond1998}. Even when the dichotomy is made clear, however, operationalizations of these languages often result in compiling action programs to logic programs \cite{Gebser2010,Ferraris2012a}, returning again to `snapshot-handling' solutions. 

The motivation behind this work stems from the hypothesis that leaving process analysis to procedural descriptions should be in principle a better choice: procedural components can directly map to native computational mechanisms, that can be used not only to \emph{re-present}, but also \emph{re-create} the process object, transforming the question from what the referent \emph{should be} (characteristic of logic), to what \textit{it is} (characteristic of simulation and more in general of model-execution). 

The paper reports therefore on a simple benchmark experiment with an \textit{hybrid} notation (that is, including procedural and declarative knowledge components), called \textit{Logic Programming Petri Nets} (LPPNs).\footnote{A prototype of a LPPN interpreter is available on \url{http://github.com/s1l3n0/pypneu}, together with the code run for conducting the experiment.} Section \ref{sec:LPPN} will introduce the motivation and an informal semantics of LPPNs. Section \ref{sec:formalization} will present a formalisation of a propositional version of LPPN. Section \ref{sec:semantics} will present an hybrid operational semantics and a denotational semantics based on ASP programs with Event Calculus. Section \ref{sec:results} will present the results of a first empirical experiment. Discussion and further developments end the paper.

\begin{figure*}[t]
	\centering
	\begin{subfigure}[b]{0.3\textwidth}
		\centering 
		\begin{tikzpicture}[font=\sffamily,node distance=1.3cm,>=stealth',shorten >=1pt,bend angle=45,auto,scale=.9]
		\tikzstyle{place}=[circle,drop shadow={opacity=.25, shadow xshift=0.07, shadow yshift=-0.07},draw=black!100,fill=white!20,minimum size=3.0mm]
		\tikzstyle{transition}=[rectangle,drop shadow={opacity=.25, shadow xshift=0.07, shadow yshift=-0.07},draw=black!100,fill=white!20,minimum size=4.0mm]
		
		\tikzstyle{every label}=[font=\sffamily\footnotesize,align=center,black]
		\begin{scope}
		\node	[place,tokens=1]	(pl1)	[label={below:\texttt{p1}}]	 at (0, 0)	{};
		\node	[place]	(pl2)	[label={below:\texttt{p3}}]	  at (2, 0)	{};
		\node	[place]	(pl3)	[label={below:\texttt{p2}}]	 at (0, -1)	{};
		\node	[transition]	(tr1)	[label={above:\texttt{t1}}]	 at (1, 0)	{}
		edge	[pre]	(pl1)
		edge	[pre]	(pl3)
		edge	[post]	(pl2);
		
		\end{scope}
		
		\end{tikzpicture} 		
		\caption{not enabled transition, before firing}
	\end{subfigure}
	~ 		\begin{subfigure}[b]{0.3\textwidth}
		\centering 
		\begin{tikzpicture}[font=\sffamily,node distance=1.3cm,>=stealth',shorten >=1pt,bend angle=45,auto,scale=.9]
		\tikzstyle{place}=[circle,drop shadow={opacity=.25, shadow xshift=0.07, shadow yshift=-0.07},draw=black!100,fill=white!20,minimum size=3.0mm]
		\tikzstyle{transition}=[rectangle,drop shadow={opacity=.25, shadow xshift=0.07, shadow yshift=-0.07},draw=black!100,fill=white!20,minimum size=4.0mm]
		
		\tikzstyle{every label}=[font=\footnotesize,align=center,black]
		\begin{scope}
		\node	[place,tokens=1]	(pl1)	[label={below:\texttt{p1}}]	 at (0, 0)	{};
		\node	[place,tokens=1]	(pl3)	[label={below:\texttt{p2}}]	 at (0, -1)	{};
		\node	[place]	(pl2)	[label={below:\texttt{p3}}]	  at (2, 0)	{};
		\node   at (1.05,1) {\LARGE\Lightning};
		\node	[transition]	(tr1)	[label={above:\texttt{t1}}]	 at (1, 0)	{}
		edge	[pre]	(pl1)
		edge	[pre]	(pl3)
		edge	[post]	(pl2);
		\end{scope}
		
		\end{tikzpicture}
		
		\caption{enabled transition and firing}
	\end{subfigure}
	~ 		\begin{subfigure}[b]{0.3\textwidth}
		\centering 
		\begin{tikzpicture}[font=\sffamily,node distance=1.3cm,>=stealth',shorten >=1pt,bend angle=45,auto,scale=.9]
		\tikzstyle{place}=[circle,drop shadow={opacity=.25, shadow xshift=0.07, shadow yshift=-0.07},draw=black!100,fill=white!20,minimum size=3.0mm]
		\tikzstyle{transition}=[rectangle,drop shadow={opacity=.25, shadow xshift=0.07, shadow yshift=-0.07},draw=black!100,fill=white!20,minimum size=4.0mm]
		
		\tikzstyle{every label}=[font=\footnotesize,align=center,black]
		\begin{scope}
		\node	[place]	(pl1)	[label={below:\texttt{p1}}]	 at (0, 0)	{};
		\node	[place]	(pl3)	[label={below:\texttt{p2}}]	 at (0, -1)	{};
		\node	[place,tokens=1]	(pl2)	[label={below:\texttt{p3}}]	  at (2, 0)	{}; 
		\node	[transition]	(tr1)	[label={above:\texttt{t1}}]	 at (1, 0)	{}
		edge	[pre]	(pl1)
		edge	[pre]	(pl3)
		edge	[post]	(pl2);
		\end{scope}
		
		\end{tikzpicture}
		
		\caption{the transition has fired\\}
	\end{subfigure}
	\caption{Example of a Petri net and of its execution (but also of a LPPN \emph{procedural component} when labels are propositions).\label{fig:lppn}}
	\vspace{-10pt}
\end{figure*}

\section{Logic Programming Petri Nets}\label{sec:LPPN}

Logic Programming Petri Nets (LPPNs) are a visual notation first introduced in \cite{Sileno2016} as an common representational ground where to align representations of \emph{law} (norms), of  \emph{implementations of law} (regulatory services in the form of business processes), and of \emph{action} (behavioural scripts ascribed to social participants). It has been used for a wide class of models (business processes embedded with normative positions, representation of scenarios issued from narratives, agent scripts, deontic paradoxes, etc. \cite{Sileno2016}). The notation builds upon the intuition that places and transitions mirror the common-sense distinction between \emph{objects} and \emph{events} (e.g. \cite{Breuker2004}), roughly reflecting the use of \emph{noun/verb} categories in language \cite{Kemmerer2010}: the procedural components captured by Petri nets can be used to model \emph{transient} aspects of the system in focus; the declarative components captured by logic programming construct can be used to model \emph{steady state} aspects, i.e. those on which the transient is irrelevant or does not make sense (e.g. terminological constraints). In this section we will informally describe the bases motivating their integration. 

\subsection{From Petri Nets to LPPNs}

Petri nets are a simple, yet effective computational modelling representation featuring an intuitive visualisation (see Fig.~1). They consist in directed, bipartite graphs with two types of nodes: \textit{places} (visually represented with circles) and \textit{transitions} (with boxes). A place can be connected only to transitions and vice-versa. One or more \emph{tokens} (dots) can reside in each place. The execution of Petri nets is also named ``{token game}'': transitions \emph{fire} by consuming tokens from their input places and producing tokens in their output places.\footnote{For an overview on the general properties of Petri nets see e.g. \cite{Murata1989}.}

Despite their widespread use in computer science, electronics, business process modelling and biology, Petri nets are generally considered not to be enough expressive for reasoning purposes: in their simplest form, tokens are indistinct, and do not transport any data. Nevertheless, it is a common practice for modellers to introduce labels to set up a correspondence between the \emph{modelling} entities and the \emph{modelled} entities. This practice enables them to read the results of a model execution in reference to the modelled system. In other words, an adequate labelling is still \emph{functional to the use} of a Petri net for modelling purposes, although it is not a requirement for the execution in itself. Further interaction is possible if these labels are processed according to an additional formalism, as for instance it occurs with \emph{Coloured Petri Nets} (CPNs) \cite{Jensen1996} (for many aspects a descendant of \emph{Predicate/Transition} nets \cite{Genrich1987}). If their expressiveness and wide application provide reasons for its adoption,  CPNs also introduce many details which are unimportant in a case modelling setting (e.g. expressions on arcs); more importantly, they still lack the ability of specifying and processing \textit{declarative bindings}, necessary, for instance, to model terminological relationships. This brings us to the idea of LPPN.

\begin{figure*}[t]
	\centering
	\begin{subfigure}[b]{0.4\textwidth}
		\centering 
		\begin{tikzpicture}[font=\sffamily,node distance=1.3cm,>=stealth',shorten >=1pt,bend angle=45,auto,scale=.9]
		\tikzstyle{place}=[circle,drop shadow={opacity=.25, shadow xshift=0.07, shadow yshift=-0.07},draw=black!100,fill=white!20,minimum size=3.0mm]
		\tikzstyle{transition}=[rectangle,draw=black!100,fill=black!100,minimum size=1.0mm]
		
		\tikzstyle{every label}=[font=\footnotesize,align=center,black]
		\begin{scope}
		\node	[place]	(pl1)	[label={left:\texttt{p4}}]	 at (0, 0)	{};
		\node	[place,tokens=1]	(pl2)	[label={left:\texttt{p5}}]	 at (0, -1)	{};
		\node	[place]	(pl3)	at (2, 0)	{};
		\node	[place]	(pl4)	[label={right:\texttt{p6}}]	 at (4, 0)	{};
		\node	[transition]	(tr1)	 at (1, 0)	[label={[yshift=-1pt]above:\texttt{AND}}]	{}
		edge	[pre]	(pl2)
		edge	[pre]   (pl1)
		edge	[post]	(pl3);
		\node	[transition]	(tr2)	 at (3, 0)	[label={[yshift=-1pt]above:\texttt{IMPLIES}}]	{}
		edge	[pre]	(pl3)
		edge	[post]	(pl4);
		\end{scope}
		
		\end{tikzpicture}
		
		\caption{\label{fig:logicLPPN}}
	\end{subfigure}
	\begin{subfigure}[b]{0.4\textwidth}
		\centering 
		\begin{tikzpicture}[font=\sffamily,node distance=1.3cm,>=stealth',shorten >=1pt,bend angle=45,auto,scale=.9]
		\tikzstyle{place}=[circle,drop shadow={opacity=.25, shadow xshift=0.07, shadow yshift=-0.07},draw=black!100,fill=white!20,minimum size=3.0mm]
		\tikzstyle{transition}=[rectangle,drop shadow={opacity=.25, shadow xshift=0.07, shadow yshift=-0.07},draw=black!100,fill=white!20,minimum size=4.0mm]
		
		\tikzstyle{every label}=[font=\footnotesize\sffamily,align=center,black]
		\tikzstyle{input}=[fill=black!20]
		\tikzstyle{output}=[fill=black!50]
		\tikzstyle{tblack}=[rectangle,draw=black!100,fill=black!100,minimum size=1.0mm]
		\tikzstyle{pblack}=[circle,draw=black!100,fill=black!100,scale=0.8]
		\begin{scope}
		\node	[place,tokens=1]	(pl1)	[label={above:\texttt{p7}}]	 at (0, 0)	{};
		\node	[place]	(pl2b)	[label={\texttt{p9}}]	 at (2, 1)	{};
		\node	[place,tokens=1]	(pl2)	[label={\texttt{p8}}]	 at (2, -2)	{};
		\node	[pblack]	(pl3)	[label={}]	 at (2, 0)	{};
		\node	[place]	(pl4)	[label={\texttt{p10}}]	 at (4, 0)	{};
		\node	[pblack]	(pl3b)	[label={}]	 at (2, -1)	{};
		\node	[place]	(pl4b)	[label={\texttt{p11}}]	 at (4, -1)	{};
		\node	[transition]	(tr1)	 at (1, 0)	[label={below:\texttt{t2}}]	{}
		edge	[pre]	(pl1)
		edge	[post]	(pl3b)
		edge	[post]	(pl3);
		\node   at (1.05,0.45) {\large\Lightning};      
		\node	[transition]	(tr2)	 at (3, 0)	[label={\texttt{t3}}]	{}
		edge	[pre]	(pl2b)    
		edge	[pre]	(pl3)
		edge	[post]	(pl4);
		\node   at (3.05,-0.55) {\large\Lightning};            
		\node	[transition]	(tr2b)	 at (3, -1)	[label={below:\texttt{t4}}]	{}
		edge	[pre]	(pl2)
		edge	[pre]	(pl3b)
		edge	[post]	(pl4b);      
		\end{scope}
		
		\end{tikzpicture} 	
		\caption{\label{fig:logicLPPN2}}
	\end{subfigure}
	
	\caption{Examples of LPPN \emph{declarative components}: (a) defined on places by means of \textit{lp-nodes} (the example corresponds to the Prolog/ASP code: \texttt{p6 :- p4, p5. p5.}); (b) defined on transitions, by means of \textit{lt-nodes}, instantaneously propagating the firing where possible (the \texttt{IMPLIES} label is here left implicit).}
	\vspace{-10pt}	
\end{figure*}

Whereas Petri nets essentially specify \emph{procedural mechanisms}, LPPNs extend those (a) with \emph{literals} as labels, attached on places and transitions; (b) with nodes specifying (logic-based) \emph{declarative bindings} on places and on transitions. This paper will focus only on propositional labelling. Under this constraint, the execution of the LPPN procedural component is the same as \emph{Condition/Event nets}, Petri nets whose places do not contain more than one token (Fig.~\ref{fig:lppn}). 

Logic operator nodes might apply on places (\textit{lp-nodes}) or on transitions (\textit{lt-nodes}). An example of a sub-net with lp-nodes (small black squares) is given in Fig.~\ref{fig:logicLPPN}; these are used to create logic compositions of places (via operators as \texttt{NEG}, \texttt{AND}, \texttt{OR}, etc). or to specify logic inter-dependencies (via the logic conditional \texttt{IMPLIES}). Similarly, transitions may be connected declaratively via lt-nodes (black circles), as in Fig.~\ref{fig:logicLPPN2}; these connections may be interpreted as channels enabling \emph{instantaneous propagation} of firing. In this case, it is not relevant to introduce operators as \texttt{AND}, for the interleaving semantics, only one source transition may fire per step. 
Operationally, the declarative components are treated integrating the \emph{stable model semantics} used in \emph{answer set programming} (ASP) \cite{Lifschitz2008b}. This was a natural choice because process execution exhibits a prototypical `forward' nature, and ASP solvers can be interpreted as providing forward chaining.\footnote{Both SLD/SLDNF resolution (Prolog) and DPLL (ASP) are based on backward chaining. In DPLL, however, all variables are grounded, and all intermediate atoms generated in the search are collected in \emph{stable models}; without defining any goal, all the worlds that are implied by the input knowledge are returned as output. From an external perspective, this is the same result we would associate with forward chaining. The intuition that there is a relation between ASP and forward chaining is confirmed e.g. in ASPeRiX \cite{Lef2009}.}

\section{Formalization\label{sec:formalization}}

This section presents a simplified version of the LPPN notation considering only a \textit{propositional} labeling. We start from the definition of propositional literals derived from ASP \cite{Lifschitz2008b}, accounting for strong and default negation.	

\begin{definition}[Literal and Extended literals] Given a set of propositional atoms $A$, the set of \emph{literals} $L = L^{+} \cup L^{-}$ consists of \emph{positive literals} (atoms) $L^{+} = A$, \emph{negative literals} (negated atoms) $L^{-} = \{ -a \; |\; a \in A\}$,	where `$-$' stands for \emph{strong negation}.\footnote{Strong negation is used to reify an explicitly false situation (e.g. ``It does not rain'').} The set of \emph{extended literals}  $L^{*} = L \cup L^{\mynot}$ consists of literals and \emph{default negation literals} $L^{\mynot} = \{ \mynot l |\; l \in L \}$, where `$\mynot$' stands for \emph{default negation}.\footnote{Default negation is used to reify a situation in which something cannot be retrieved/inferred (e.g. `It is unknown whether it rains or not').}
\end{definition}

\noindent We denote the basic topology of a Petri net as a procedural net.

\begin{definition}[Procedural net]A \emph{procedural net} is a \emph{bipartite directed graph} connecting two finite sets of nodes, called \emph{places} and \emph{transitions}. It can be written as $ N = \tuple{P, T, E}$, where $P = \{p_{1}, \ldots, p_{n}\}$ is the set of \emph{place} nodes; $T = \{t_{1}, \ldots, t_{m}\}$ is the set of \emph{transition} nodes; $E = E^+ \cup E^-$ is the set of \emph{arcs} connecting them: $E^+$ from transitions to places, $E^{-}$ from places to transitions.
\end{definition}

\noindent LPPNs consists of three components: a procedural net specifying causal relationships, and two declarative nets specifying respectively logical dependencies at the level of objects or ongoing events (on places), and on impulse events (on transitions). Furthermore, \emph{propositional} LPPNs build upon a boolean marking on places (like \emph{condition/event} nets).


\begin{definition}[Propositional Logic Programming Petri Net] A \emph{propositional Logic Programming Petri Net} $\mathit{LPPN}_\mathrm{prop}$ is a Petri Net whose places and transitions are labeled with literals, enriched with declarative nets of places and of transitions. It is defined by the following components:
	\begin{itemize} 
		\item $\tuple{P, T, \mathit{PE}}$ is a procedural net; $\mathit{PE}$ stands for \emph{procedural edges};
		\item $C_{P} : P \rightarrow L^{*}$ and $C_{T} : T \rightarrow L$ are labeling functions, associating literals respectively to places and to transitions;	
		
		\item $\mathit{OP} = \{\neg, -, \wedge, \vee, \rightarrow, \leftrightarrow, \ldots \}$ is a set of logic operators.	
		\item $\mathit{LP}$ and $\mathit{LT}$ are 
		sets of \emph{logic operator nodes} respectively for places (\emph{lp-nodes}) and for transitions (\emph{lt-nodes}).
		\item $C_{\mathit{LP}} : \mathit{LP} \rightarrow \mathit{OP}$ maps each lp-node to a logic operator; similarly,  $C_{\mathit{LT}} : \mathit{LT} \rightarrow \mathit{OP}$ does the same for lt-nodes.
		\item $\mathit{DE}_{\mathit{LP}} = \mathit{DE}^+_{\mathit{LP}} \cup \mathit{DE}^-_{\mathit{LP}}$
		is the set of arcs connecting lp-nodes to places; similarly,  $\mathit{DE}_{\mathit{LT}} = \mathit{DE}^+_{\mathit{LT}} \cup \mathit{DE}^-_{\mathit{LT}}$ for lt-nodes and transitions.\footnote{Note that  $\mathit{DE}^-_{\mathit{LT}} \subseteq (T \cup P) \times \mathit{LT}$, i.e. these edges go from transitions \emph{and} places (modeling contextual conditions) to lt-nodes.}
		
							
		\item $M: P \rightarrow \{0, 1\}$ returns the marking of a place, i.e. whether the place contains (1) or does not contain (0) a token.
	\end{itemize}
\end{definition}

\noindent Note that if $\mathit{LP} \cup \mathit{LT} = \varnothing$, we have a \emph{strictly procedural} $\mathit{LPPN}_\mathrm{prop}$, i.e. a standard binary Petri net. If $T = \varnothing$, we have a \emph{strictly declarative} $\mathit{LPPN}_\mathrm{prop}$, that can be directly mapped to an ASP program.

%

For simplicity, we overlook in this document the syntaxic constraints on the network topology which are inherited from ASP. 

\section{Semantics\label{sec:semantics}}

This section will present two semantics for LPPNs: a hybrid operational semantics and a denotational semantics, based on ASP and event calculus.

\subsection{Hybrid operational semantics}

The execution cycle of a LPPN consists of four steps: 
\begin{enumerate}
\item given a ``source'' marking $M$, the bindings of the declarative net of places entail a ``ground'' marking $M^*$; 
\item an enabled transition is selected to \emph{pre-fire}, so determining a ``source'' \emph{transition-event} $e$; 
\item the bindings of the declarative net of transitions entail all propagations of this event, obtaining a set of \emph{transition-events}, also denoted as the ``ground'' \emph{event-marking} $E^*$; 
\item all transition-events are fired, producing and consuming the relative tokens.
\end{enumerate}
The steps (1) and (3) are processed in distinct moments by an ASP solver: the declarative nets of places (1) or of transitions (3) are translated as \emph{rules}, tokens (1) or transition-events (3) are reified as \emph{facts}. The ASP solver takes as input the resulting program and, if satisfiable, it provides as output respectively one or more ground marking (1) or one or more sets transition-events to be fired (3). 

The steps (2) and (4) make clear the distinction the \emph{external} firings (the ``source'' transition-event selected at execution level) from the \emph{internal} firing, immediately propagated (the ``ground'' transition-events triggered by the declarative net of transitions). 
The following definitions provides a formalisation of these concepts.


\begin{definition}[Enabled transition]
	A transition $t$ is \emph{enabled} in a ground marking $M^*$ if a token is available for each input places: $$
	\mathit{Enabled}(t) \equiv \forall p_i \in \bullet t, M^*(p) = 1 $$
\end{definition}
Similarly to what marking is for places, we consider an \emph{event-marking} for transitions $E: T\rightarrow \{0, 1\}$. $E(t) = 1$ if the transition $t$ produces a transition-event $e$. Each step $s$ has a ``source'' event-marking $E$.
\begin{definition}[Pre-firing] An enabled transition $t$ \emph{pre-fires} (implicitly, at a step $s$) if it is selected to produce a \emph{transition-event}: $$
	\forall t \in \mathit{Enabled}(T) :
	t \; \textrm{pre-fires} \equiv 
	E(t) = 1$$
\end{definition}
Applying an \emph{interleaving semantics} for the pre-firing, the interpreter selects only one transition to pre-fire per step; for any other $t'$, $E(t') = 0$.

\begin{definition}[Firing] An enabled transition $t$ \emph{fires} (implicitly, at a step $s$) by propagation consuming a token from each input place, and producing a token in each output place:
	\begin{equation*}
	\begin{gathered}
	\forall t \in \mathit{Enabled}(T) :
	t \; \textrm{fires} \equiv \\
	E^*(t) = 1 \leftrightarrow 
	\forall p_i \in \bullet t : M'(p_i) = 0 \; \wedge \;
	\forall p_o \in t \bullet: M'(p_o) = 1 
	\end{gathered}
	\end{equation*}
\end{definition}

\begin{figure}[t]
	\centering
		\begin{tikzpicture}[font=\sffamily,node distance=1.3cm,>=stealth',shorten >=1pt,bend angle=45,auto,scale=.9]
\tikzstyle{place}=[circle,drop shadow={opacity=.25, shadow xshift=0.07, shadow yshift=-0.07},draw=black!100,fill=white!20,minimum size=3.0mm]
\tikzstyle{transition}=[rectangle,drop shadow={opacity=.25, shadow xshift=0.07, shadow yshift=-0.07},draw=black!100,fill=white!20,minimum size=4.0mm]

\tikzstyle{every label}=[font=\footnotesize\sffamily,align=center,black]
\tikzstyle{input}=[fill=black!20]
\tikzstyle{output}=[fill=black!50]
\tikzstyle{tblack}=[rectangle,draw=black!100,fill=black!100,minimum size=1.0mm]
\tikzstyle{pblack}=[circle,draw=black!100,fill=black!100,scale=0.8]
\begin{scope}

\node	[place,tokens=1]	(pl1)	[label={below:\texttt{c1}}]	 at (0, 0)	{};
\node	[place]	(pl2)	[label={\texttt{c3}}]	 at (2, -2)	{};
\node	[place]	(pl3)	[label={below:\texttt{c2}}]	 at (2, 0)	{};
\node	[place]	(pl4)	[label={below:\texttt{c5}}]	 at (4, 0)	{};
\node	[pblack]	(pl3b)	[label={right:\texttt{IMPLIES}}]	 at (1, -1)	{};
\node	[place,tokens=1]	(pl4b)	[label={\texttt{c4}}]	 at (0, -2)	{};
\node	[transition]	(tr0)	 at (-1, 0)	[label={above:\texttt{e2}}]	{}
edge	[pre]	(pl1);
\node	[transition]	(tr1)	 at (1, 0)	[label={above:\texttt{e1}}]	{}
edge	[pre]	(pl1)
edge	[post]	(pl3b)
edge	[post]	(pl3);
\node	[tblack]	(tr2)	 at (3, 0)	[label={above:\texttt{IMPLIES}}]	{}
edge	[pre]	(pl3)
edge	[post]	(pl4);
\node	[transition]	(tr2b)	 at (1, -2)	[label={below:\texttt{e3}}]	{}
edge	[post]	(pl2)
edge	[pre]	(pl3b)
edge	[pre]	(pl4b);      
\end{scope}
	\end{tikzpicture}
\caption{Example of LPPN with procedural and declarative components.\label{fig:fulllppn}}
\end{figure}

\paragraph{Running example} Let us consider the LPPN in Fig.~\ref{fig:fulllppn}. Here, for simplicity, we will conflate the names of the transition/places with their labels; in the general case this should be made different as there might be multiple nodes with the same label. The proposed net specifies causal mechanisms, declarative constraints. There is only one token in \texttt{c1}, enabling the transitions associated to \texttt{e1} and \texttt{e2}. The following execution paths are possible: (1) \texttt{e2} fires, consuming the token in \texttt{c1}, \texttt{e3} fires, consuming the token \texttt{c4} and producing a token in \texttt{c3}; (2, 3) \texttt{e3} fires, and then one of \texttt{e1} or \texttt{e2} fires; (4) \texttt{e1} fires, consuming the token in \texttt{c1}; the firing propagates to \texttt{e3}; the source firing of \texttt{e1} also produces a token in \texttt{c2}; the existence of \texttt{c2} is a sufficient condition for immediately reifying \texttt{c5}.

\subsection{Denotational semantics}

One of the possibilities to validate a formal language is to map it into another formal language, i.e. to provide a \emph{denotational semantics}. The declarative component of a LPPN, by design, can be directly rewritten as ASP code. As we are already halfway down the path, we can translate the remaining procedural component into ASP.

\subsubsection{Event Calculus axioms}

A well-known solution to treat change in logic programming is \emph{event calculus} (EC) \cite{Kowalski1986,Shanahan1999}. The simple version is already satisfactory for our purposes. A modification of the original axioms is necessary to deal with the \emph{locality} brought by places and transitions:

\begin{lstlisting}
holdsAt(F, P, N) :-
   initially(F, P), not clipped(0, F, P, N),
   fluent(F), place(P), time(N).

holdsAt(F, P, N2) :-
   firesAt(T, N1), N1 < N2,
   initiates(T, F, P, N1), not clipped(N1, F, P, N2),
   place(P), transition(T), fluent(F), time(N1), time(N2).

clipped(N1, F, P, N2) :-
   firesAt(T, N), N1 <= N, N < N2, 
   terminates(T, F, P, N),
   place(P), transition(T), fluent(F), time(N1), time(N2), time(N).
\end{lstlisting}

\subsubsection{Interleaved semantics axioms}The interleaved semantics can be translated into the following rules:
\renewcommand{\theenumi}{\roman{enumi}} 
\begin{enumerate}
	\item all enabled transitions may or may not pre-fire;
	\item pre-firing is transformed to firing;
	\item at least one enabled transition must pre-fire per step, i.e. it is impossible that no transition fire if there are enabled transitions; 
	\item at maximum one transition can pre-fire per step.
\end{enumerate}
\noindent In ASP code:
\begin{lstlisting}
{prefiresAt(T, N)} :-                                    % (i)
   enabled(T, N), transition(T), time(N).

firesAt(e1, N) :- prefiresAt(e1, N).                     % (ii)

someTransitionPrefiresAt(N) :-                           % (iii)
   prefiresAt(T, N), transition(T), time(N).
:- not someTransitionPrefiresAt(N), enabled(T, N), transition(T), time(N).

:- prefiresAt(T1, N), prefiresAt(T2, N), T1 != T2,       % (iv)
   transition(T1), transition(T2), time(N).
\end{lstlisting}

\subsubsection{Transformation of a LPPN to an ASP program\label{sec:mappingtoASP}}
The mapping of a given LPPN to an equivalent ASP program includes the previous axioms and the output of the following steps:

\renewcommand{\theenumi}{\roman{enumi}} 
\begin{enumerate}
	\item for each place $p$, with label $C_P(p)$ 
	\begin{enumerate}
		\item type it as place,
		\item specify its initial state,
		\item for each place with more than one output, write down that you cannot consume more than the only available token.
	\end{enumerate}
	\item for each transition $t$, with label $C_T(t)$
	\begin{enumerate}
		\item type it as transition, 
		\item define the conditions for which it is enabled,
		\item for each output place, define how to create tokens in the output places,
		\item for each input place, define how to consume tokens in the output places.
	\end{enumerate}
	\item for each lp-node $\mathit{lp}$,
	\begin{enumerate}
		\item specify the logic constraint to be applied between inputs and outputs.
	\end{enumerate}
	\item for each lt-node $\mathit{lt}$,
	\begin{enumerate}
		\item write down the logic dependencies between transitions allowing the propagation.
	\end{enumerate}
\end{enumerate}

\noindent As a concrete example, we apply these actions on some of the components of the LPPN in Fig.~\ref{fig:fulllppn}:

\begin{lstlisting}
fluent(filled).

%%% p1, associated to c1
place(c1).                                                          % 1.a
initially(filled, c1).                                              % 1.b
:- 2{terminates(e2, filled, c1, N); terminates(e1, filled, c1, N)}. % 1.c

%%% t1, associated to e1
transition(e1).	                                                    % 2.a
enabled(e1, N) :- holdsAt(filled, c1, N).                           % 2.b
terminates(e1, filled, c1, N) :- firesAt(e1, N).                    % 2.c 
initiates(e1, filled, c2, N) :- firesAt(e1, N).                     % 2.d

%% lp1 
holdsAt(filled, c5, N) :- holdsAt(filled, c2, N).                   % 3.a

%% lt1
firesAt(e3, N) :- firesAt(e1, N), enabled(e1, N).                   % 4.a
              
\end{lstlisting}

\paragraph{Execution} With the transformation steps given above, valid LPPNs can be transformed into ASP programs. In particular, for the axioms presented here, these programs can be solved the ASP engine \texttt{clingo} \cite{Eiter2008a}, also available online at: \url{https://potassco.org/clingo/run/}. Setting a temporal range (with the instruction ``\texttt{time(0..$n$).}'') the output answer sets represent all possible executions path after at most $n$ steps. 

\begin{figure}[t]
	\centering
	\scalebox{.25}{
		\includegraphics{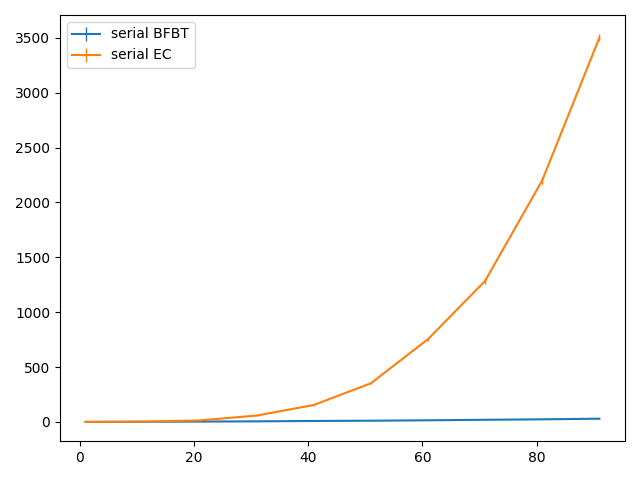}
		\includegraphics{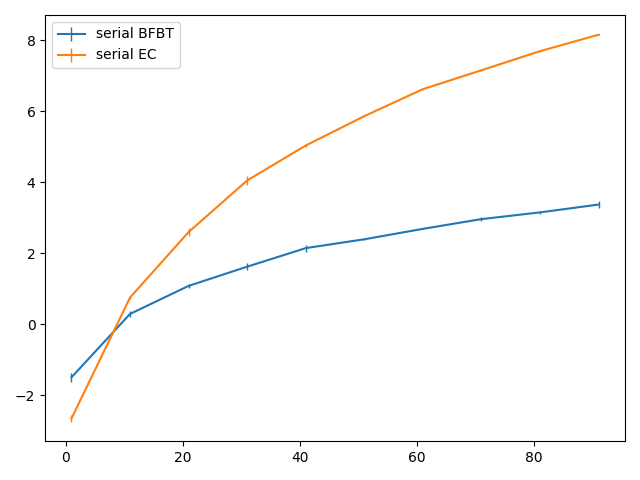}}
	\scalebox{.25}{
		\includegraphics{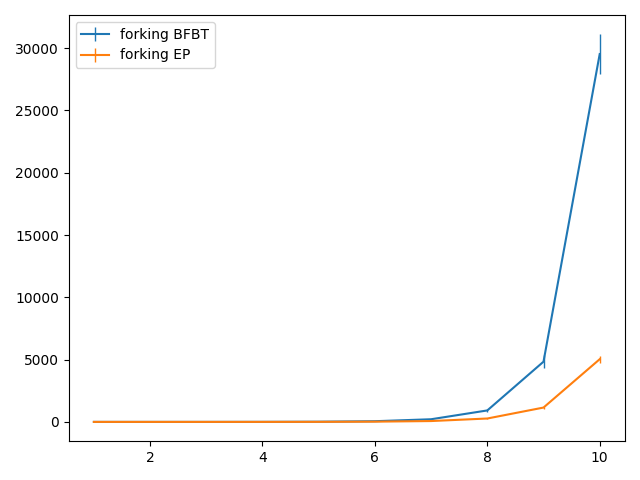}
		\includegraphics{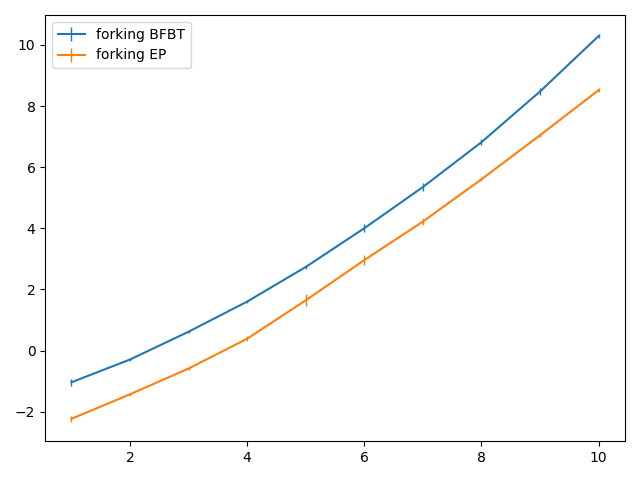}
	}
	\caption{Average execution times (ms) in linear and logarithmic scales over 10 executions of serial and forking configurations of propositional LPPNs of different depths, performed following alternatively the hybrid operational semantics, via \emph{brute force} execution and \emph{backtracking} (BF+BT); and the denotational semantics, via \emph{event calculus} (EC). Data is on Table~\ref{tab:evaluation}.\label{fig:evaluation}}
\end{figure}

\setlength{\tabcolsep}{0.5em} 
{\renewcommand{\arraystretch}{1.2}
	\begin{table}[t]
		\centering
		\scalebox{.7}{
			
		\begin{tabular}{llllll}
			& \multicolumn{5}{c}{depth of composition of minimal structures} \\ 
			\textit{serial} & 1 & 11 & 21 & 31 & 41 \\ \hline
			EC & 0.1 $\pm$ 0.0 & 2.1 $\pm$ 0.1  & 13.5 $\pm$ 1.8  & 58.0 $\pm$ 8.2  & 154.2 $\pm$ 8.6  \\
			BF+BT & 0.2 $\pm$ 0.0 & 1.3 $\pm$ 0.1 & 2.9 $\pm$ 0.2 & 5.1 $\pm$ 0.6 & 8.6 $\pm$ 0.9 \\ \hline
			 & 51 & 61 & 71 & 81 & 91 \\ \hline
			EC & 352.4 $\pm$ 9.9 & 754.4 $\pm$ 15.2  & 1285.3 $\pm$ 29.3  & 2200.6 $\pm$ 30.1  & 3499.0 $\pm$ 33.9  \\
			BF+BT & 11.0 $\pm$ 0.2 & 14.7 $\pm$ 0.2 & 19.4 $\pm$ 1.1 & 23.4 $\pm$ 0.8 & 29.3 $\pm$ 3.1 \\ \hline
			\\
			\textit{forking} & 1 & 2 & 3 & 4 & 5 \\ \hline
			EC & 0.1 $\pm$ 0.0 & 0.2 $\pm$ 0.0  & 0.6 $\pm$ 0.0  & 1.5 $\pm$ 0.1  & 5.3 $\pm$ 1.3  \\
			BF+BT & 0.4 $\pm$ 0.0 & 0.7 $\pm$ 0.0 & 1.9 $\pm$ 0.1 & 5.0 $\pm$ 0.3 & 15.5 $\pm$ 1.2 \\ \hline
			 & 6 & 7 & 8 & 9 & 10 \\ \hline
			EC & 19.6 $\pm$ 3.5 & 68.5 $\pm$ 8.2  & 272.6 $\pm$ 20.4  & 1151.3 $\pm$ 83.7  & 5033.8 $\pm$ 291.7  \\
			BF+BT & 55.3 $\pm$ 7.7 & 213.1 $\pm$ 36.0 & 920.9 $\pm$ 112.1 & 4834.5 $\pm$ 537.9 & 29529.9 $\pm$ 1665.4 \\ \hline
		
	\end{tabular}
		}
		\vspace{10pt}
		\caption{Average execution time (ms) over 10 executions of different configurations of propositional LPPNs, performed following alternatively the hybrid operational semantics, via \emph{brute force} execution and \emph{backtracking} (BF+BT); and the denotational semantics, via \emph{event calculus} (EC).\label{tab:evaluation}}
	\end{table}

\section{Results\label{sec:results}}
The proposal presented above has been used for developing a prototype Python application for specifying, executing and analyzing LPPNs\footnote{Available at \url{http://github.com/s1l3n0/pypneu}.}; it exploits \texttt{clingo} \cite{Eiter2008a}, as this provide runtime interfaces enabling a direct control of the solver instance without regrounding the program at each cycle. 
This enabled us to perform some direct evaluation of any given LPPN input.

When we process the input LPPN by means of the denotational semantics, the input is transformed to an ASP program, and the solver intervenes \emph{fully} to provide the possible execution paths. Instead, when we refer to the hybrid operational semantics, the solver intervenes only \emph{partially} in the execution cycle, to entail the constraints implied by the declarative components of the net; the rest of the computational burden is on the module responsible for the Petri net execution. In this context, one might ask if we can observe some performances between these two alternative modes of analysis/execution.

At the moment, we have only evaluated a propositional version of LPPN, and a limited series of structures, namely compositions of minimal \emph{serial} elements (a transition with an input and output places) or minimal \emph{forking} elements (a place with two output transitions). In order to implement the procedural component of the operational semantics, the current Petri Net analysis module builds upon a simple \textit{brute force} (BF) execution algorithm, and \emph{depth-first} search with \emph{backtracking} (BT) to cover all the possible execution paths. 

Table~\ref{tab:evaluation} summarises the performances of 10 executions of different network configurations.\footnote{The tests were run on a MacBook Pro (2018) provided with a 2.2 GHz 6-core processor Intel Code i7 and 16Gb RAM DDR4.} Results are also illustrated on Fig.~\ref{fig:evaluation}. The data essentially confirms our hypothesis: the analysis based on the operational semantics (BF+BT) clearly outperforms and scales excellently for the serial configurations, while that based on the denotational semantics (EC) scales poorly in this configuration. For the forking configurations, BF+BT is evidently slower in absolute terms. Intuitively this is due to the efficient search and pruning capabilities of ASP. Unlike clingo, the Python code of the Petri net executor/analyzer is not optimised; on the contrary, for many aspects this represents a lower-bound on the possible implementation choices. Nevertheless, if we consider execution times in logarithmic scale, we observe that the two methods are essentially comparable in terms of tractability.

\section{Conclusion}

The paper presents an empirical experiment with LPPNs, a logic programming-based extension to Petri Nets. LPPNs were introduced with a practical goal in mind: a visual modelling notation relatively simple for non-experts, that could handle explicit declarative knowledge, and that could model causation and other procedural aspects \cite{Sileno2016}. It was inspired by the point made in \cite{Kowalski2009} on the widespread confusion in cognitive science and computational disciplines around the notion of \emph{rules} (namely between declarative and reactive rules). Previous contributions \cite{Sileno2016,Sileno2017a} highlighted the potential use of LPPNs in normative modelling tasks in combination with business process modelling, thus potentially facilitating cross-fertilization between theoretical to practical settings. 


Here the focus has been put on its computational properties, showing that maintaining the two levels separated can bring better performances. The declarative dimension allows to treat at higher abstraction phenomena for which there is a viable specification at outcome level. The procedural dimension works better for processes that can be directly executed. 


Future developments concern the extension of this work to a wider range of experiments, first considering mixed networks (of declarative, procedural components) with mixed configurations (serial compositions, forks, joins, etc.) and then passing to the extended LPPN notation accounting for predicates. The actual impact on real models should be evaluated as well: scenarios describing cases have very few forks, they rather function as \emph{orchestrated} (i.e. directed from the scenario) \emph{scripts} (procedural models distributed amongst actors). Consequently, applications that require the use of scenarios (e.g. for interpretation, model-based diagnosis, conformance checking, etc.) may take advantage of the hybrid operational semantics. The computational improvement may be further extended considering existing proposals in the literature. For instance, execution algorithms alternative to brute execution \cite{Moreno2007,Piedrafita2011}; or decomposition techniques, for instance in \emph{single-entry-single-exit} (SESE) components \cite{Munoz-Gama2014}, that open up the possibility of concurrent execution. 

Further, these results should be confronted with existing techniques for handling temporal reasoning and causality, e.g. the already cited \textit{Action languages} \cite{Gelfond1998}, related works (e.g. F2LP \cite{Lee2009a}) and applications (CCalc, Coala, Cplus2ASP); optimized versions of Event Calculus (e.g. \cite{Artikis2015}); applications based on LTL, CTL and related formalisms. 


\bibliographystyle{splncs03} 
\bibliography{CAUSAL.0}

\begin{thebibliography}{10}
\providecommand{\url}[1]{\texttt{#1}}
\providecommand{\urlprefix}{URL }

\bibitem{Artikis2015}
Artikis, A., Sergot, M., Paliouras, G.: {An Event Calculus for Event
  Recognition}. IEEE Transactions on Knowledge and Data Engineering  27(4),
  895--908 (2015)

\bibitem{Breuker2004}
Breuker, J., Hoekstra, R.: {Core concepts of law: taking common-sense
  seriously}. Proc. of Formal Ontologies in Information  (2004)

\bibitem{Eiter2008a}
Eiter, T., Faber, W., Fink, M., Woltran, S.: {A user's guide to gringo, clasp,
  clingo, and iclingo}. Annals of Mathematics and Artificial Intelligence
  51(2-4),  123--165 (2008)

\bibitem{Ferraris2012a}
Ferraris, P., Lee, J.: {Representing first-order causal theories by logic
  programs}. Theory and Practice of Logic Programming  12(03),  383--412 (may
  2012)

\bibitem{Gebser2010}
Gebser, M., Grote, T., Schaub, T.: {Coala: A compiler from action languages to
  ASP}. Lecture Notes in Computer Science (including subseries Lecture Notes in
  Artificial Intelligence and Lecture Notes in Bioinformatics)  6341 LNAI,
  360--364 (2010)

\bibitem{Gelfond1998}
Gelfond, M., Lifschitz, V.: {Action languages}. Electronic Transactions on AI
  (1998)

\bibitem{Genrich1987}
Genrich, H.J.: {Predicate/Transition Nets}. In: Proceedings Advances in Petri
  nets 1986. pp. 207--247 (1987)

\bibitem{Jensen1996}
Jensen, K.: {Coloured Petri Nets: Basic Concepts, Analysis Methods and
  Practical Use}. Springer-Verlag, London, UK (1996)

\bibitem{Kemmerer2010}
Kemmerer, D., Eggleston, A.: {Nouns and verbs in the brain: Implications of
  linguistic typology for cognitive neuroscience}. Lingua  120(12),  2686--2690
  (2010)

\bibitem{Kowalski2009}
Kowalski, R., Sadri, F.: {Integrating logic programming and production systems
  in abductive logic programming agents}. Web Reasoning and Rule Systems  LNCS
  5837,  1--23 (2009)

\bibitem{Kowalski1986}
Kowalski, R., Sergot, M.: {A logic based calculus of events}. New Generation
  Computing  4(June 1975),  67--95 (1986)

\bibitem{Lee2009a}
Lee, J., Palla, R.: {System f2lp - computing answer sets of first-order
  formulas}. Lecture Notes in Computer Science (including subseries Lecture
  Notes in Artificial Intelligence and Lecture Notes in Bioinformatics)  5753
  LNAI,  515--521 (2009)

\bibitem{Lef2009}
Lefevre, C., Nicolas, P.: {A First Order Forward Chaining for Answer Set
  Computing}. LPNMR 2009  LNCS 5753,  196--208 (2009)

\bibitem{Lifschitz2008b}
Lifschitz, V.: {What Is Answer Set Programming?} Proceedings of the AAAI
  Conference on Artificial Intelligence  (2008)

\bibitem{McCarthy1969}
McCarthy, J., Hayes, P.J.: {Some philosophical problems from the standpoint of
  artificial intelligence}. In: Machine Intelligence, pp. 1--51. Edimburgh
  University Press (1969)

\bibitem{Moreno2007}
Moreno, R.P., Salcedo, J.L.V.: {Performance evaluation of petri nets execution
  algorithms}. Conference Proceedings - IEEE International Conference on
  Systems, Man and Cybernetics pp. 1400--1407 (2007)

\bibitem{Munoz-Gama2014}
Munoz-Gama, J., Carmona, J., {Van Der Aalst}, W.M.P.: {Single-Entry Single-Exit
  decomposed conformance checking}. Information Systems  46,  102--122 (2014)

\bibitem{Murata1989}
Murata, T.: {Petri nets: Properties, analysis and applications}. Proceedings of
  the IEEE  77(4) (1989)

\bibitem{Piedrafita2011}
Piedrafita, R., Villarroel, J.L.: {Performance evaluation of petri nets
  centralized implementation. The execution time controller}. Discrete Event
  Dynamic Systems: Theory and Applications  21(2),  139--169 (2011)

\bibitem{Reiter2001}
Reiter, R.: {Knowledge in action: logical foundations for specifying and
  implementing dynamical systems}. MIT Press (2001)

\bibitem{Shanahan1999}
Shanahan, M.: {The event calculus explained}. Artificial Intelligence Today pp.
  409--430 (1999)

\bibitem{Sileno2016}
Sileno, G.: {Aligning Law and Action}. Ph.D. thesis, University of Amsterdam
  (2016)

\bibitem{Sileno2017a}
Sileno, G., Boer, A., van Engers, T.: {A Petri net-based notation for normative
  modeling: evaluation on deontic paradoxes}. In: Workshop on MIning and
  REasoning with Legal texts (MIREL2017) in conjunction with ICAIL2017 (2017)

\bibitem{Thielscher1999}
Thielscher, M.: {From situation calculus to fluent calculus: State update
  axioms as a solution to the inferential frame problem}. Artificial
  Intelligence  111(1-2),  277--299 (1999)

\end{thebibliography}

\end{document}